\documentclass{article}

\usepackage{microtype}
\usepackage{graphicx}
\usepackage{subfigure}
\usepackage{booktabs} % for professional tables
\usepackage{listings}
\usepackage{xcolor}
\usepackage{appendix}
\usepackage{hyperref}

\lstdefinestyle{jsonstyle}{
  language=json,
  basicstyle=\ttfamily\footnotesize,
  backgroundcolor=\color{white},
  keywordstyle=\color{blue},
  stringstyle=\color{red},
  commentstyle=\color{green},
  frame=single,
  breaklines=true,
  columns=flexible
}

\usepackage[accepted]{icml2024}

\usepackage{amsmath}
\usepackage{amssymb}
\usepackage{mathtools}
\usepackage{amsthm}

\usepackage[capitalize,noabbrev]{cleveref}

\theoremstyle{plain}

\theoremstyle{definition}

\theoremstyle{remark}

\usepackage[textsize=tiny]{todonotes}

\icmltitlerunning{LegalBench-RAG: A Benchmark for Retrieval-Augmented Generation in the Legal Domain}

\begin{document}
\twocolumn[
\icmltitle{LegalBench-RAG: A Benchmark for Retrieval-Augmented Generation in the Legal Domain}

% It is OKAY to include author information, even for blind
% submissions: the style file will automatically remove it for you
% unless you've provided the [accepted] option to the icml2024
% package.

% List of affiliations: The first argument should be a (short)
% identifier you will use later to specify author affiliations
% Academic affiliations should list Department, University, City, Region, Country
% Industry affiliations should list Company, City, Region, Country

% You can specify symbols, otherwise they are numbered in order.
% Ideally, you should not use this facility. Affiliations will be numbered
% in order of appearance and this is the preferred way.
\icmlsetsymbol{equal}{*}

\begin{icmlauthorlist}
\icmlauthor{Nicholas Pipitone}{equal,yyy}
\icmlauthor{Ghita Houir Alami}{equal,yyy}
\end{icmlauthorlist}

\icmlaffiliation{yyy}{ZeroEntropy, San Francisco, CA}

\icmlcorrespondingauthor{Nicholas Pipitone}{npip99@gmail.com}
\icmlcorrespondingauthor{Ghita Houir Alami}{ghita.houir-alami@polytechnique.edu}

% You may provide any keywords that you
% find helpful for describing your paper; these are used to populate
% the "keywords" metadata in the PDF but will not be shown in the document
\icmlkeywords{Machine Learning, ICML}

\vskip 0.3in
]

% this must go after the closing bracket ] following \twocolumn[ ...

% This command actually creates the footnote in the first column
% listing the affiliations and the copyright notice.
% The command takes one argument, which is text to display at the start of the footnote.
% The \icmlEqualContribution command is standard text for equal contribution.
% Remove it (just {}) if you do not need this facility.

%\printAffiliationsAndNotice{}  % leave blank if no need to mention equal contribution
\printAffiliationsAndNotice{\icmlEqualContribution} % otherwise use the standard text.

\begin{abstract}
Retrieval-Augmented Generation (RAG) systems are showing promising potential, and are becoming increasingly relevant in AI-powered legal applications. Existing benchmarks, such as LegalBench, assess the generative capabilities of Large Language Models (LLMs) in the legal domain, but there is a critical gap in evaluating the retrieval component of RAG systems.
To address this, we introduce \textbf{LegalBench-RAG}, the first benchmark specifically designed to evaluate the retrieval step of RAG pipelines within the legal space. LegalBench-RAG emphasizes precise retrieval by focusing on extracting minimal, highly relevant text segments from legal documents. These highly relevant snippets are preferred over retrieving document IDs, or large sequences of imprecise chunks, both of which can exceed context window limitations. Long context windows cost more to process, induce higher latency, and lead LLMs to forget or hallucinate information. Additionally, precise results allow LLMs to generate citations for the end user. The LegalBench-RAG benchmark is constructed by retracing the context used in LegalBench queries back to their original locations within the legal corpus, resulting in a dataset of 6,858 query-answer pairs over a corpus of over 79M characters, entirely human-annotated by legal experts.
We also introduce LegalBench-RAG-mini, a lightweight version for rapid iteration and experimentation.
By providing a dedicated benchmark for legal retrieval, LegalBench-RAG serves as a critical tool for companies and researchers focused on enhancing the accuracy and performance of RAG systems in the legal domain. The LegalBench-RAG dataset is publicly available at \href{https://github.com/zeroentropy-cc/legalbenchrag}{https://github.com/zeroentropy-cc/legalbenchrag}
\end{abstract}

\begin{figure}[ht]
\vskip 0.2in
\begin{center}
\centerline{\includegraphics[width=\columnwidth]{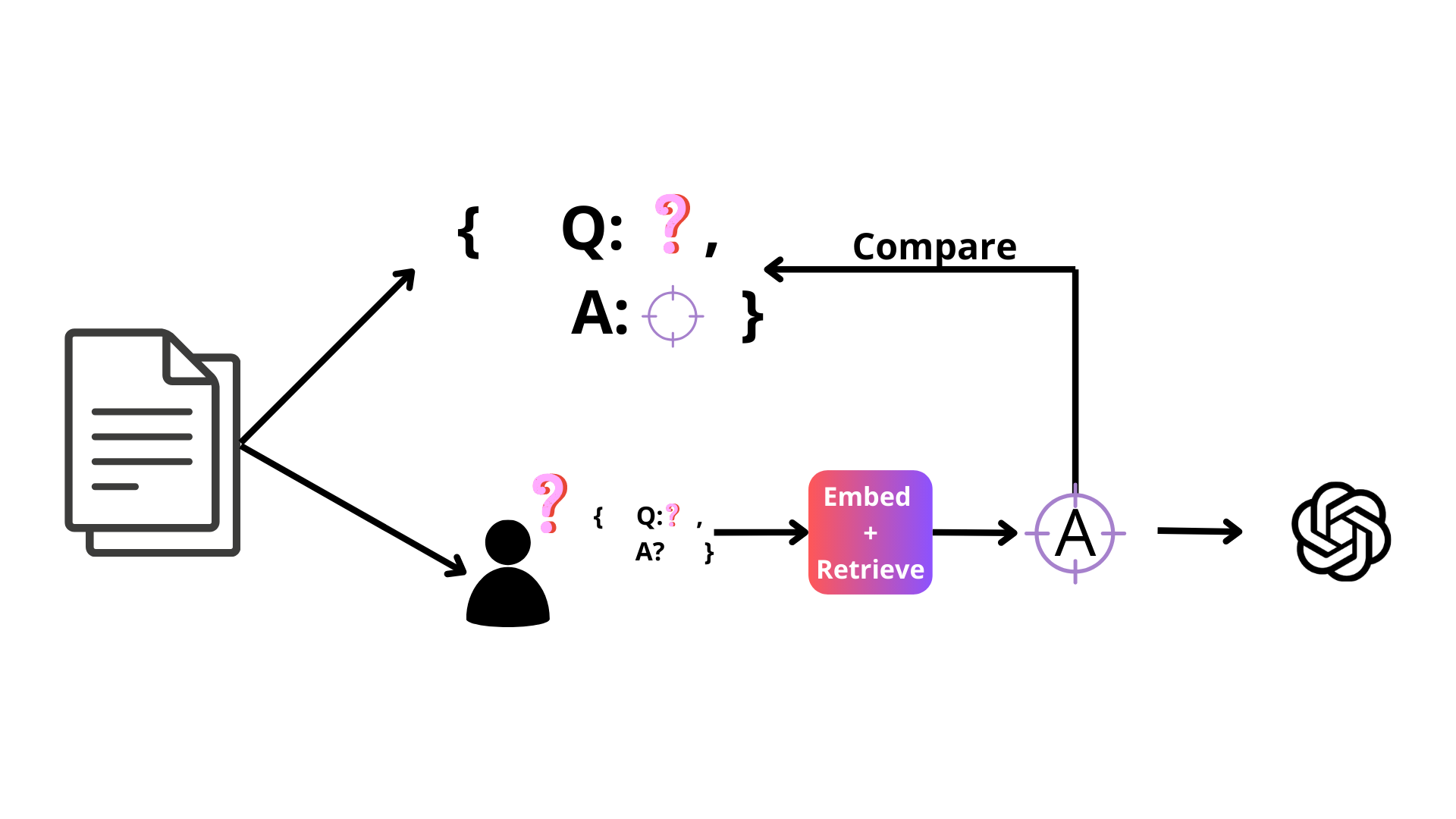}}
\caption{Benchmarking The Retrieval Step Of RAG Systems}
\label{icml-historical}
\end{center}
\vskip -0.2in
\end{figure}

\section{Introduction}
\label{sec}

In the rapidly evolving landscape of AI in the legal sector, Retrieval-Augmented Generation (RAG) \cite{DBLP:journals/corr/abs-2005-11401} systems have emerged as a crucial technology. These systems, which combine retrieval mechanisms with generative large language models (LLMs), show promising potential for contextualized generation. However, as companies race to develop RAG-based solutions, a critical gap in the ecosystem remains unaddressed: the lack of a dedicated benchmark for evaluating the retrieval component in legal-specific RAG systems.
Existing benchmarks such as LegalBench \cite{guha2023legalbenchcollaborativelybuiltbenchmark} assess the reasoning capabilities of LLMs on complex legal questions, and how effectively LLMs recall legal knowledge in their training set. However, these existing benchmarks do not evaluate the retrieval quality over a large corpus, which is crucial for RAG-based systems. Similarly, a legal-specific RAG benchmark is essential because legal documents have unique structures, terminologies, and requirements that more general RAG benchmarks cannot adequately assess. General benchmarks often lack the domain-specific nuances and complex legal relationships found in real-world use cases. These benchmarks \cite{yu2024evaluationretrievalaugmentedgenerationsurvey} also generally focus on recall without distinguishing between the retrieval of large, contextually broad chunks and the precise retrieval of small, highly relevant text snippets. The risk of hallucination at generation increases as irrelevant information is included in the context window of an LLM, which poses a significant risk in the legal industry. Benchmarks such as RAGTruth \cite{niu2024ragtruthhallucinationcorpusdeveloping} study the extent of hallucinated content at the generation step of RAG systems. Additionally, succinct annotations into highly relevant text snippets allow a human-in-the-loop to quickly verify the veracity of an LLM's claims.
To address all of these issues, a benchmark is necessary to compare the quality of competing retrieval algorithms. For this, we introduce LegalBench-RAG, the first benchmark designed specifically for evaluating retrieval systems in the legal domain. LegalBench-RAG provides a rigorous framework to assess how well retrieval mechanisms can pinpoint exact legal references, offering a more granular and relevant measure of performance than existing benchmarks.

\begin{table*}[h!]
\centering
\begin{tabular}{@{}lcccc@{}}
\hline
\textbf{Q\&A Count} & \textbf{Dataset Name} & \textbf{Dataset Description}\\
\hline
946 & ContractNLI & NDA related documents \\
4042 & Contract Understanding Atticus Dataset (CUAD) & Private Contracts\\
1676 & M\&A Understanding Dataset & M\&A documents of public companies \\
194 & Privacy QA & Privacy policies of consumer apps \\
\hline
\textbf{6858} & 4 datasets & Variety of legal documents\\
\hline
\end{tabular}
\caption{Summary of Datasets}
\label{tab:datasets}
\end{table*}

\section{Related Work}

\subsection{Retrieval Augmented Generation (RAG)}

A Retrieval-Augmented Generation (RAG) \cite{DBLP:journals/corr/abs-2005-11401} system is an intelligent generative system that utilizes a knowledge base, denoted as $\mathcal{D}$, which contains a set of documents. In this case, each document, represented as $d_i \in  \mathcal{D}$, is a string of legal text. A traditional implementation will segment each document $d_i$ into a set of chunks $c_j \in \mathcal{C}_i$. These chunks are then transformed into vector embeddings using a specialized embedding model.
After the ingestion of the documents, a user can submit a query $q$, which will be vectorized using the same embedding model. The system then retrieves the top-k chunks ${\mathcal{R}_q} = \{r_1, r_2, \ldots, r_K\}$ most relevant to the query, using similarity metrics such as cosine similarity.
The retrieved chunks, together with the query and a system prompt $P$, are processed by a large language model (LLM) to generate the final response. This overall process is formalized as:

 $$\text{Contextual Retriever}(\mathbf{q}, \mathcal{D}) \rightarrow {\mathcal{R}_\mathbf{q}}$$
 $$
 \text{LLM}_P(\mathbf{q},{\mathcal{R}_\mathbf{q}} ) \rightarrow  \text{answer}$$

 \textbf{Contextual Retriever}: The contextual retriever module locates relevant information from an external knowledge repository, returning a corresponding context set ${\mathcal{R}_q}$. Typically, RAG architectures incorporate a bi-encoder retriever such as DPR \cite{karpukhin2020densepassageretrievalopendomain}, known for its efficiency and accuracy in information retrieval.
 The Contextual Retriever module often includes a reranking step. First, the top-\(k'\) items are retrieved using a bi-encoder retriever. Then, all $k'$ items will be reranked using a cross-encoder model that outputs a similarity score between $q$ and each item. The top $k < k'$ results are returned.\\
\textbf{Answer Generator}: The generator component, which often leverages a sequence-to-sequence model, receives both the question and the context as inputs. It then generates an answer \( y_{j,\mathbf{q}} \) with the likelihood \( P_G(y_{j,\mathbf{q}} \mid \mathbf{q}, r_{j,\mathbf{q}}) \). 

 \subsection{Retrieval Augmented Generation (RAG) Benchmarks}

Question-Answer Benchmarks have been introduced prior to the adoption of RAG systems, and  datasets such as HotPotQA \cite{yang2018hotpotqadatasetdiverseexplainable} are still in use today. Multiple RAG-specific benchmarks have also been created to assess the quality of RAG systems. Retrieval Augmented Generation Benchmark (RGB) \cite{chen2023benchmarkinglargelanguagemodels} and RECALL \cite{liu2023recallbenchmarkllmsrobustness} are two major contributions that asses the performance of RAG models. However, these benchmarks evaluate a simple case where the answer of a query can be retrieved and solved in a general context. More complex datasets such as MultiHop-RAG \cite{tang2024multihopragbenchmarkingretrievalaugmentedgeneration} were also introduced to assess the retrieval and reasoning capability of LLMs for complex multi-hop queries.

\subsection{LegalBench}

LegalBench \cite{guha2023legalbenchcollaborativelybuiltbenchmark} was introduced to enable greater study of legal reasoning capabilities of LLMs. It is a collaboratively constructed legal reasoning benchmark consisting of 162 tasks covering six different types of legal reasoning. LegalBench was built through an interdisciplinary process, in which tasks designed and hand-crafted by legal professionals were collected. Because these subject matter experts took a leading role in construction, tasks either measure legal reasoning capabilities that are practically useful, or measure reasoning skills that lawyers find interesting. 

\subsection{Other legal focused benchmarks}

In recent years, there has been considerable research aimed at evaluating AI models' abilities to undertake tasks traditionally handled by legal professionals \cite{maroudas2022legaltechopendiarieslesson} \cite{katz2023naturallanguageprocessinglegal}.

Initial studies focused on complex legal tasks, such as document review \cite{wang2023maudexpertannotatedlegalnlp} and case summarization \cite{shen2022multilexsumrealworldsummariescivil} \cite{shukla2022legalcasedocumentsummarization}.

Subsequent research efforts have been directed towards the challenges posed by legal texts, such as extensive document lengths and specialized jargon \cite{10.1145/3594536.3595142}, \cite{chalkidis-etal-2022-lexglue}.

Another critical area of investigation has been the creation of tasks that assess various forms of inferential reasoning commonly required in legal contexts \cite{chalkidis-etal-2022-lexglue} \cite{guha2023legalbenchcollaborativelybuiltbenchmark}.

\section{A Benchmarking Dataset: LegalBench-RAG}

\subsection{LegalBench-RAG Constuction}
\label{final author}

In this section, we provide detailed information on the construction of the LegalBench-RAG dataset. Notably, we describe the process of creating a set of Queries along with the corresponding ground truth snippets sets derived from the well-known LegalBench \cite{guha2023legalbenchcollaborativelybuiltbenchmark} dataset. 

\subsubsection{Starting Point: LegalBench}
Our benchmark development is based on LegalBench, a collaboratively constructed legal reasoning benchmark consisting of 162 tasks covering six different types of legal reasoning. LegalBench is a widely recognized dataset that tests LLMs' understanding of legal concepts and their legal reasoning by providing them with the exact context needed to answer legal questions. LegalBench focuses solely on evaluating the generation phase of the RAG pipeline, assessing how well the LLM can generate accurate responses given a certain context, task, and prompt. However, LegalBench does not benchmark the ability to extract the correct context from within a larger corpus. This limitation is what inspired the LegalBench-RAG benchmark.

We selected four datasets to construct our retrieval benchmark: Privacy Question Answering (PrivacyQA) \cite{ravichander2019questionansweringprivacypolicies}, Contract Understanding Atticus Dataset (CUAD) \cite{hendrycks2021cuadexpertannotatednlpdataset}, Mergers and Acquisitions Understanding Dataset (MAUD) \cite{wang2023maudexpertannotatedlegalnlp} and Contract Natural Language Inference (ContractNLI) \cite{koreeda2021contractnlidatasetdocumentlevelnatural}. 

\subsubsection{Tracing Back to Original Sources}
To transform LegalBench into a retrieval benchmark, we undertook a comprehensive process to trace each text segment used in LegalBench, back to its original location within the source corpus. Each of the four source datasets used is unique, but in general, the datasets have categorical annotations and associated context clauses for each annotation. Our desired output is a pairing between queries and relevant spans, where each span is a range of characters in the original corpus. Creating this dataset involved searching in the original corpus for the context clauses, in order to deduce the index span implied. And, it involved transforming annotations into queries. By creating these pairings between queries and lists of relevant spans, we ensured that the benchmark accurately reflects the retrieval capabilities needed to locate exactly the relevant information within a large legal corpus. This detailed index mapping is critical for evaluating retrieval-based models within the legal domain.

\subsubsection{Construction Process}

Each of the four source datasets is converted into LegalBench-RAG queries through a slightly different process, but all four follow a similar formula.

First, as a pre-processing step, we must create unique descriptions of every document in the corpus, and we must create a mapping between annotation categories and interrogatives.

From there, we can now generate our queries. A full LegalBench-RAG query is constructed using the format:
\begin{quote}
\texttt{"Consider \textit{(document\_description)}; \textit{(interrogative)}"}
\end{quote}
Therefore, every query is a combination of the document description, and an interrogative. Each individual query in LegalBench-RAG originates from an individual annotation in the source dataset. The annotation itself provides the document ID and annotation category, which we convert into a description and interrogative using our pre-processed mappings.

\subsubsection{Example Annotations}
For instance, consider the Contract Understanding Atticus Dataset (CUAD) "Affiliate License-Licensee" task from LegalBench. 

The goal of this task is to classify if a clause describes a license grant to a licensee (incl. sublicensor) and the affiliates of such licensee/sublicensor.

In LegalBench, a query appears as follows:

\noindent\hrulefill
\begin{quote}
    \textit{\textbf{Query}: Does the clause describe a license grant to a licensee (incl. sublicensor) and the affiliates of such licensee/sublicensor?} \\
    \textit{\textbf{Clause}: Supplier hereby grants Bank of America a nonexclusive, (...) restrictions of this Section. \\
    \textbf{Label}: Yes} \\
    
    \textit{\textbf{Query}: Does the clause describe a license grant to a licensee (incl. sublicensor) and the affiliates of such licensee/sublicensor?} \\
    \textit{\textbf{Clause}: Promptly upon receipt of notice from Pfizer, Exact and Pfizer shall engage in exclusive good faith negotiations to enter into a definitive written agreement for the Ex-US Commercial Rights.\\
    \textbf{Label}: No}
\end{quote}
\noindent\hrulefill

In the original CUAD dataset, annotations appear as follows:

\noindent\hrulefill
\begin{quote}
    \textit{\textbf{Clause}: Supplier hereby grants Bank of America a nonexclusive, (...) restrictions of this Section. \\
    \textbf{File}: \normalfont{\texttt{CardlyticsInc\_20180112\_S-1\_EX-\\10.16\_11002987\_EX-10.16\_Maintenance Agreement1.pdf}}\\}
    \textbf{Label}: Irrevocable Or Perpetual License\\

    \textit{\textbf{Clause}: Promptly upon receipt of notice from Pfizer, Exact and Pfizer shall engage in exclusive good faith negotiations to enter into a definitive written agreement for the Ex-US Commercial Rights. \\
    \textbf{File}: \normalfont{\texttt{ExactSciencesCorp\_20180822\_8-K\\\_EX-10.1\_11331629\_EX-10.1\_Promotion Agreement.pdf}}\\}
    \textbf{Label}: Rofr/Rofo/Rofn\\
\end{quote}
\noindent\hrulefill

In our LegalBench-RAG dataset, queries appear as follows:

\noindent\hrulefill
\begin{quote}
    \textit{\textbf{Query}: Consider the Software License, Customization, and Maintenance Agreement between Cardlytics, Inc. and Bank of America; Are the licenses granted under this contract non-transferable?\\
    \textbf{Label (File)}: \normalfont{\texttt{CardlyticsInc\_20180112\_S-1\\\_EX-10.16\_11002987\_EX-10.16\_Maintenance Agreement1.txt}}\\
    \textbf{\textit{Label (Span)}}: \normalfont{\texttt{[44579, 45211]}}} \\
    
    \textit{\textbf{Query}: Consider the Cologuard Promotion Agreement between Exact Sciences Corporation and Pfizer Inc.; Does this contract include any right of first refusal, right of first offer, or right of first negotiation?\\
    \textbf{Label (File)}: \normalfont{\texttt{ExactSciencesCorp\_20180822\_8\\-K\_EX-10.1\_11331629\_EX-10.1\_Promotion Agreement.txt}}\\
    \textbf{\textit{Label (Span)}}: \normalfont{\texttt{[51520, 52804]}}}
\end{quote}
\noindent\hrulefill

Notice the distinguishing characteristics between LegalBench and LegalBench-RAG. LegalBench will take a given query, and convert the annotation label from the CUAD dataset, into a Yes/No label based on whether or not the label matches the query. LegalBench-RAG on the other hand, will ask the same query, and the label will be the filename and span of the relevant text necessary to answer to the query.

A detail omitted for brevity, is that LegalBench-RAG labels will be not be a single filename and index range, but rather an array of (filename, index range) tuples, as the source dataset has annotations that are sets of non-adjacent spans.

\subsection{Quality Control}

Quality control is of crucial importance for any RAG benchmark, to ensure trust in recall and precision scores. All of the annotations themselves were created by domain-experts via the methods described by our four source datasets. On top of this, our team conducted a thorough manual inspection of every data point in the dataset, employing the process described below. Quality control was rigorously applied at three critical decision points:

\textbf{Mapping Annotation Categories to Interrogatives } \\
The first step involved creating a mapping from annotation categories to interrogatives. Unlike the original LegalBench, where questions could be directly utilized, LegalBench-RAG necessitates that all text relevant to the query is included in the annotation, while irrelevant text is excluded. To meet these stringent criteria, we manually constructed a mapping where the relevant text would align precisely with the domain experts’ annotations. Categories with inconsistent annotation precision were excluded entirely from our dataset.

\textbf{Mapping Document IDs to Descriptions} \\
Second, we created the mapping from document IDs to document descriptions. For this, we utilized \textit{GPT-4o-mini} to automatically create a short description for each document, utilizing the file name and approximate first and last paragraphs of the document. For consistency, the model was given explicit instructions on how to format its description, and regular expressions (regex) was used to validate the output format. Each description was then manually inspected. We also used embedding similarity on every pair of document descriptions to find pairs of most similar descriptions, manually excluding any pairs that could not be distinctly differentiated.

\textbf{Selection of Annotation Categories} \\
The final decision point was determining which annotation categories to include. This decision was informed by our manual evaluations of the precision levels across categories, with inconsistent categories being excluded to preserve the integrity of the benchmark.

Through these quality control measures, we have verified the robustness and reliability of the LegalBench-RAG dataset, ensuring that each critical decision point aligns with the intended benchmark standards.

\subsection{Dataset Structure}

The LegalBench-RAG benchmark is structured around two primary components: the original corpus and the QA pairs.

The corpus includes the documents from our four source datasets. It excludes documents that were not asked by at least one query in LegalBench-RAG. Additionally, every document is in .txt format, utilizing the txt files provided by the source datasets. We opt to utilize the original uncleaned, often messy, file names representative of file names found in the industry.

The QA pairs are directly linked to the documents within the corpus. \hyperref[sec:appendixA]{Appendix A} provides a detailed structure of these QA pairs. Each query is associated with a list of relevant snippets extracted from the different documents in the corpus that directly answer the query. For each snippet, the file path, the exact quote, and the precise character indices within the document are provided, ensuring a clear reference to the original source. This detailed mapping is crucial for accurately assessing retrieval capabilities within the LegalBench-RAG benchmark.

\subsubsection{Descriptive Statistics}

LegalBench-RAG is composed of 4 datasets and totals almost 80 million characters in its corpus across 714 documents. Each pair is annotated by legal experts, ensuring the highest accuracy and relevance for this benchmark. We extract 6,889 question-answer pairs which constitutes our retrieval benchmark. The outcome is a robust dataset that we call LegalBench-RAG. The statistics regarding this corpus are presented in Table \ref{tab:dataset_stats}.

\subsubsection{LegalBench-RAG and LegalBench-RAG-mini}

Given the consequent size of LegalBench-RAG, this paper also introduces LegalBench-RAG-mini, a more lightweight version of the benchmark we proposed. LegalBench-RAG-mini was created by selecting exactly 194 queries from each of the four datasets PrivacyQA, CUAD, MAUD and ContractNLI. We select the portions of the corpus corresponding to these queries accordingly. This results in a dataset of 776 queries as described in Table \ref{tab:mini}.

\begin{table*}[]
\centering
\begin{tabular}{@{}lccccc@{}}
\toprule
\textbf{ } & \textbf{ContractNLI} & \textbf{MAUD} & \textbf{CUAD} & \textbf{PrivacyQA}& \textbf{Total} \\ 
\midrule
\textbf{Number of Documents} & 95 & 150 & 462 & 7 & 714\\
\textbf{Number of Corpus Characters} & 1,013,969 & 52,721,337 & 25,792,044 & 176,864 & 79,704,214 \\
\textbf{Average Document Length} & 10,673 & 351,476 & 55,827 & 25,266 & 443,242\\
\textbf{Number of Queries} & 977 & 1,676 & 4,042 & 194& 6,889\\
\textbf{Link} & \href{https://aclanthology.org/2021.findings-emnlp.164/}{ContractNLI} & \href{https://www.atticusprojectai.org/maud}{MAUD} & \href{https://www.atticusprojectai.org/cuad}{CUAD} & \href{https://arxiv.org/abs/1911.00841}{PrivacyQA} \\ \bottomrule
\end{tabular}
\caption{Dataset Statistics}
\label{tab:dataset_stats}
\end{table*}

\begin{table*}[]
\centering
\begin{tabular}{@{}lccccc@{}}
\hline
\textbf{ }                 & \textbf{ContractNLI} & \textbf{MAUD} & \textbf{CUAD} & \textbf{PrivacyQA}& \textbf{Total} \\
\hline
\textbf{Number of Documents}               & 18       & 18       & 29       & 7& 72  \\
\hline
\textbf{Number of Corpus Characters} & 184,267 & 7,109,200 & 1,211,773 & 176,864 & 8,682,104 \\
\hline
\textbf{Number of Queries}                & 194      & 194      & 194      & 194 & 776 \\
\hline
\end{tabular}
\caption{Summary of datasets in LegalBench-RAG-mini, including the number of documents, corpus characters, and queries.}
\label{tab:mini}
\end{table*}

\subsection{Significance of this work}

Creating datasets in the legal space is a difficult, time-consuming and costly task. For instance, The CUAD dataset alone, was a year-long effort that required the expertise of over 40 lawyers to generate 13,000 annotations. Given that 9,283 pages were reviewed at least 4 times, that each page required 5-10 minutes of review, and assuming a rate of \$500 per hour for legal expertise, then we can estimate that the creation of the CUAD dataset would cost around \$2,000,000 to replicate.

LegalBench-RAG is the first publicly available retrieval-focused legal benchmark, and can be used for both commercial and academic purposes to assess the quality of the retrieval step of RAG pipelines on legal tasks. We hope that the introduction of this dataset allows the industry to more easily compare and iterate upon the plethora of RAG techniques available today, by providing a standardized evaluation framework for these techniques.

\subsection{Limitations}
\label{sec:limitations}

We make note of the several limitations of this dataset. Our four source datasets include NDAs, M\&A agreements, various commercial contracts, and the privacy policies of consumer-facing online companies. While this is broad, this dataset is not made of an exhaustive list of all existing documents in the legal industry. For example, this benchmark does not assess structured numerical data parsing, which is relevant for cases involving financial fraud. It also does not assess the parsing and analyzing of medical records, which is relevant in personal injury suits.

Notably, the queries in this benchmark are always answered by exactly one document, so this benchmark does not assess the ability of a retrieval system to reason across information found in multiple documents. It only assesses the ability of a retrieval system to select the correct document, and then the correct snippets within that document.

Through manual inspection, we were able to find several queries that do require multi-hop reasoning, which is a difficult task to achieve. However, there is room to create even more complexr queries that require a very high number of hops to generate the correct snippets.

\section{Benchmarking RAG systems using LegalBench-RAG}

A typical RAG system operates in two phases: the retrieval phase and the generation phase. LegalBench-RAG is designed to assess the effectiveness of the retrieval phase, while the original LegalBench can be utilized to evaluate the performance of the generation phase. In this section, we demonstrate how LegalBench-RAG can be employed to measure the quality of retrieval. 
The code to run the benchmark is publicly available  \href{https://github.com/zeroentropy-cc/legalbenchrag}{here}.\\
All the experiments conducted in this work are run on LegalBench-RAG-mini.

\begin{table*}[h!]
\centering
\begin{tabular}{lccccccc|ccccccc}
\toprule
& \multicolumn{7}{c}{Precision @ $k$} & \multicolumn{7}{c}{Recall @ $k$} \\
\cmidrule(lr){2-8} \cmidrule(lr){9-15}
& 1 & 2 & 4 & 8 & 16 & 32 & 64 & 1 & 2 & 4 & 8 & 16 & 32 & 64 \\
\midrule
\textbf{PrivacyQA}  & 7.86 & 7.31 & 6.41 & 5.06 & 3.58 & 2.41 & 1.54 & 7.45 & 12.53 & 20.88 & 32.38 & 42.45 & 54.27 & 66.07\\
\textbf{ContractNLI}  & \textbf{16.45} & \textbf{14.80} & \textbf{12.53} & \textbf{9.73} & \textbf{6.70} & \textbf{4.65} & \textbf{3.04} & 11.32 & 19.10 & \textbf{29.79} & \textbf{45.59} & \textbf{56.75} & \textbf{69.88} & \textbf{86.57}\\
\textbf{MAUD} & 3.36 & 2.65 & 2.18 & 1.89 & 1.48 & 1.06 & 0.75 & 2.54 & 3.12 & 4.53 & 8.75 & 13.16 & 18.36 & 25.62\\
\textbf{CUAD} & 9.27 & 8.05 & 5.98 & 4.33 & 2.77 & 1.77 & 1.09 & \textbf{12.60} & \textbf{19.47} & 27.92 & 40.70 & 51.02 & 64.38 & 75.71\\
\textbf{ALL}  & 2.40 & 3.76 & 4.97 & 4.33 & 3.39 & 2.17 & 1.29 & 3.37 & 8.44 & 21.30 & 34.51 & 48.88 & 64.47 & 76.39\\
\bottomrule
\end{tabular}
\caption{Performance comparison on different datasets for Precision and Recall at various $k$ values for the Naive Method.}
\label{tab:chunking_performance_naive}
\end{table*}

\begin{table*}[h!]
\centering
\begin{tabular}{lccccccc|ccccccc}
\toprule
& \multicolumn{7}{c}{Precision @ $k$} & \multicolumn{7}{c}{Recall @ $k$} \\
\cmidrule(lr){2-8} \cmidrule(lr){9-15}
& 1 & 2 & 4 & 8 & 16 & 32 & 64 & 1 & 2 & 4 & 8 & 16 & 32 & 64 \\
\midrule

\textbf{Privacy QA}  & \textbf{14.38} & \textbf{13.55} & \textbf{12.34} & \textbf{9.03} & \textbf{6.06} & \textbf{4.17} & \textbf{2.81}& \textbf{8.85} & \textbf{15.21} & \textbf{27.92} & \textbf{42.37} & \textbf{55.12} & \textbf{71.19} & \textbf{84.19}\\
\textbf{ContractNLI}  & 6.63 & 5.29 & 3.89 & 2.81 & 1.98 & 1.29 & 0.90 & 7.63 & 11.33 & 17.34 & 24.99 & 35.80 & 46.57 & 61.72\\
\textbf{MAUD}  & 2.65 & 1.77& 1.96 & 1.40 & 1.39 & 1.15 & 0.82 & 1.65 & 2.09 & 4.59 & 6.18 & 12.93 & 21.04 & 28.28\\
\textbf{CUAD}  & 1.97 & 4.03 & 4.83& 4.20 & 2.94 & 1.99 & 1.25 & 1.62 & 8.11 & 17.72 & 31.68 & 44.38 & 60.04 & 74.70\\
\textbf{ALL}  & 6.41 & 6.16 & 5.76& 4.36 & 3.09 & 2.15 & 1.45 & 4.94 & 9.19 & 16.90 & 26.30 & 37.06& 49.71 & 62.22\\

\bottomrule
\end{tabular}
\caption{Performance comparison on different datasets for Precision and Recall at various $k$ values for the Recursive Character Text Splitter Method.}
\label{tab:chunking_performance_rcts}
\end{table*}

\subsection{Hyperparameters in RAG Pipelines}

Implementing a RAG pipeline involves selecting various hyperparameters and making critical design decisions, which can be adjusted to optimize performance. In this section, we discuss a few design decisions that will be evaluated using LegalBench-RAG.

\subsubsection{Pre-Processing Strategies}

As discussed in Section 2.1, standard RAG pipelines typically generate embeddings from document chunks using an embedding model. The method by which documents are divided into these chunks is a crucial design consideration \cite{pinecone2024chunking}. Different strategies can be employed, ranging from simple fixed-size chunks with potentially overlapping segments, to recursive and contextual approaches, to more sophisticated semantic chunking methods \cite{retrievaltutorials2024}. The choice of chunking strategy can significantly impact the effectiveness of the retrieval system.

\subsubsection{Post-Processing Strategies}

Following retrieval, several critical design decisions must be made in the post-processing phase. One key parameter is the number of retrieved chunks (i.e., the choice of $k$ in top-k retrieval) that are subsequently input to the large language model (LLM). This selection is pivotal due to the inherent trade-off between maximizing the relevant context provided to the LLM and minimizing the introduction of noise, which could increase the risk of hallucination. Furthermore, the decision to employ a reranker model on the retrieved chunks represents another crucial consideration, as it can substantially alter the final set of chunks presented to the LLM, potentially impacting the model’s overall performance.

\subsubsection{Other design decisions}

Among other critical decisions, the choice of the embedding model to use is a common one. Embedding models can vary significantly in how well they represent textual content. A well-chosen embedding model can enhance retrieval accuracy by producing more meaningful and contextually relevant embeddings, which improves the quality of the retrieval.

\subsection{Experimental Setup}

In this evaluation experiment, we implement multiple RAG pipelines that we benchmark using LegalBench-RAG. We conduct several experiments to study the impact of the chunking strategy and the effect of reranking the retrieved chunks using a specialized model.

We used OpenAI's "text-embedding-3-large" embeddings for the embedding model \cite{openai2024embeddingmodels}. We chose SQLite Vec as a vector database and Cohere's "rerank-english-v3.0" as our reranker model \cite{cohere2024rerank}. The two chunking strategies we benchmarked here are (1) a naive fixed-size chunking method with a chunk size of 500 characters with no overlap, and (2) a Recursive Character Text Splitter (RCTS) \cite{langchain2024recursivetextsplitter}, which divides text into smaller chunks by sequentially attempting to split on a predefined list of characters which maintains paragraphs, sentences, and words together.
The two post-processing we evaluated here are (1) no reranker after the retrieval, and (2) Cohere Reranker after the retrieval.

To evaluate the results on LegalBench-RAG, we weight the metrics equally on each dataset, independently from the number of documents or queries they contain.

\subsection{Variation of preprocessing splitting strategies}

In our initial experiment, we fixed all hyperparameters and compared the impact of transitioning from a naive chunking strategy to a Recursive Text Character Splitter (RTCS). The performance results for each dataset, as well as varying values of $k$, are presented in Tables \ref{tab:chunking_performance_naive} and \ref{tab:chunking_performance_rcts}.

\begin{table*}[h!]
\centering
\begin{tabular}{lccccccc|ccccccc}
\toprule
& \multicolumn{7}{c}{Precision @ $k$} & \multicolumn{7}{c}{Recall @ $k$} \\
\cmidrule(lr){2-8} \cmidrule(lr){9-15}
& 1 & 2 & 4 & 8 & 16 & 32 & 64 & 1 & 2 & 4 & 8 & 16 & 32 & 64 \\
\midrule
\textbf{PrivacyQA} & \textbf{14.38} & \textbf{13.55} & \textbf{12.34} & \textbf{9.02} & \textbf{6.06} & \textbf{4.17} & \textbf{2.81} & \textbf{8.85} & \textbf{15.21} & \textbf{27.92} & \textbf{42.37} & \textbf{55.12} & \textbf{71.19} & \textbf{84.19} \\
\textbf{ContractNLI} & 6.63 & 5.28 & 3.89 & 2.81 & 1.98 & 1.29 & 0.90 & 7.63 & 11.34 & 17.34 & 24.99 & 35.80 & 46.57 & 61.72 \\
\textbf{MAUD} & 2.64 & 1.77 & 1.96 & 1.40 & 1.38 & 1.15 & 0.82 & 1.65 & 2.09 & 5.59 & 6.18 & 12.93 & 21.04 & 28.28 \\
\textbf{CUAD} & 1.97 & 4.03 & 4.83 & 4.20 & 2.94 & 1.99 & 1.25 & 1.62 & 8.11 & 17.72 & 31.68 & 44.38 & 60.04 & 74.70 \\
\textbf{ALL}\textsuperscript{†} & 6.41 & 6.16 &  5.76& 4.36 & 3.09 & 2.15 & 1.45 & 4.94 & 9.19 &  16.90& 26.30 & 37.05 & 49.71 & 62.22 \\
\bottomrule
\end{tabular}
\caption{Performance comparison on different datasets for Precision and Recall at various $k$ values for the Naive Method with the Cohere Reranker.}
\label{tab:chunking_performance_reranker}
\end{table*}

\begin{table*}[h!]
\centering
\begin{tabular}{lccccccc|ccccccc}
\toprule
& \multicolumn{7}{c}{Precision @ $k$} & \multicolumn{7}{c}{Recall @ $k$} \\
\cmidrule(lr){2-8} \cmidrule(lr){9-15}
& 1 & 2 & 4 & 8 & 16 & 32 & 64 & 1 & 2 & 4 & 8 & 16 & 32 & 64 \\
\midrule
\textbf{PrivacyQA} & \textbf{13.94} & \textbf{15.91} & \textbf{13.32} & \textbf{9.57} & \textbf{6.88} & \textbf{4.68} & \textbf{3.28} & \textbf{7.32} & \textbf{16.12} & \textbf{25.65} & \textbf{35.60} & \textbf{51.87} & \textbf{64.98} & \textbf{79.61} \\
\textbf{ContractNLI} & 5.08 & 5.59 & 5.04 & 3.67 & 2.52 & 1.75 & 1.17 & 4.91 & 9.33 & 16.09 & 25.83 & 35.04 & 46.90 & 62.97\\
\textbf{MAUD} & 1.94 & 2.63 & 2.05 & 1.77 & 1.79 & 1.55 & 1.12 & 0.52 & 2.48 & 4.39 & 7.24 & 14.03 & 22.60 & 31.46  \\
\textbf{CUAD} & 3.53 & 4.18 & 6.18 & 5.06 & 3.93 & 2.74 & 1.66 & 3.17 & 7.33 & 18.26 & 28.67 & 42.50 & 55.66 & 70.19 \\
\textbf{ALL}\textsuperscript{†} & 6.13 & 7.08 & 6.65 & 5.02 & 3.78 & 2.68 & 1.81 & 3.98 & 8.82 & 16.10 & 24.34 & 35.86 & 47.54 & 61.06 \\
\bottomrule
\end{tabular}
\caption{Performance comparison on different datasets for Precision and Recall at various $k$ values for the RCTS Method with the Cohere Reranker.}
\label{tab:rcts_reranker}
\end{table*}

\subsection{Variation of postprocessing method}

We conduct a second experiment where we freeze all other hyperparameters and change the post-processing method used only. We compare using no reranking stategy of the retrieved chunks and using Cohere's Reranker model. The results of this comparison are shown in Tables \ref{tab:chunking_performance_naive} and \ref{tab:chunking_performance_reranker} respectively. We also run our benchmark on the RCTS method with Cohere's Reranker in Table \ref{tab:rcts_reranker}.

\section{Results and Discussion}

\subsection{Results of the experimentation}

\begin{figure}[H]
\centering
\vskip -0.2in
    \begin{minipage}[b]{0.45\textwidth}
        \centering
    \centerline{\includegraphics[width=\columnwidth]{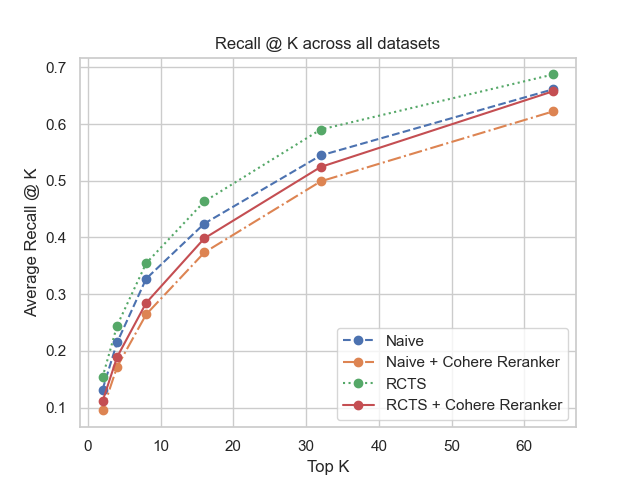}}
    \caption{Recall @ $k$ across all datasets}
    \label{fig:avgrec}
\end{minipage}\hfill
\begin{minipage}[b]{0.45\textwidth}
    \centering
    \centerline{\includegraphics[width=\columnwidth]{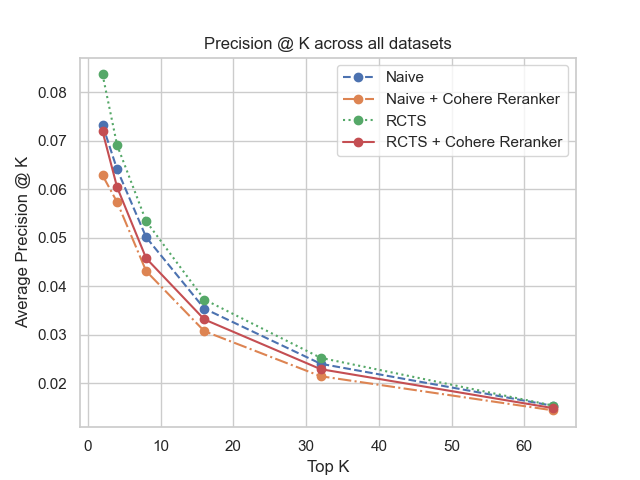}}
    \caption{Precision @ $k$ across all datasets}
    \label{fig:avgprec}
\end{minipage}
\vskip -0.2in
\end{figure}

The overall evaluation results are computed for varying values of $k$ between 1 and 64 and for each of the four dataset and are also aggregated across all datasets using an equal weight for each dataset. Figures  \ref{fig:avgrec} and \ref{fig:avgprec} display Recall@$k$ and Precision@$k$, respectively, for all method combinations.

This experiment demonstrates that the most effective strategy was the Recursive Text Character Splitter (RTCS) without a reranker. Surprisingly, the performance of the Cohere Reranker was inferior compared to not using a reranker. This result may be attributed to the difficulty of this benchmark and its focus on legal text, which may not align well with a general-purpose model like Cohere’s reranker. Both precision and recall were highest with the RTCS + No reranker configuration. As anticipated, recall improved with increasing values of  $k$, while precision decreased. The observed low precision values could be due to the highly targeted and concise nature of the ground truth.

\subsection{Comparison of the four datasets}

\begin{figure}[]
\vskip 0.1in
\centering
    \begin{minipage}[b]{0.45\textwidth}
        \centering
\centerline{\includegraphics[width=\columnwidth]{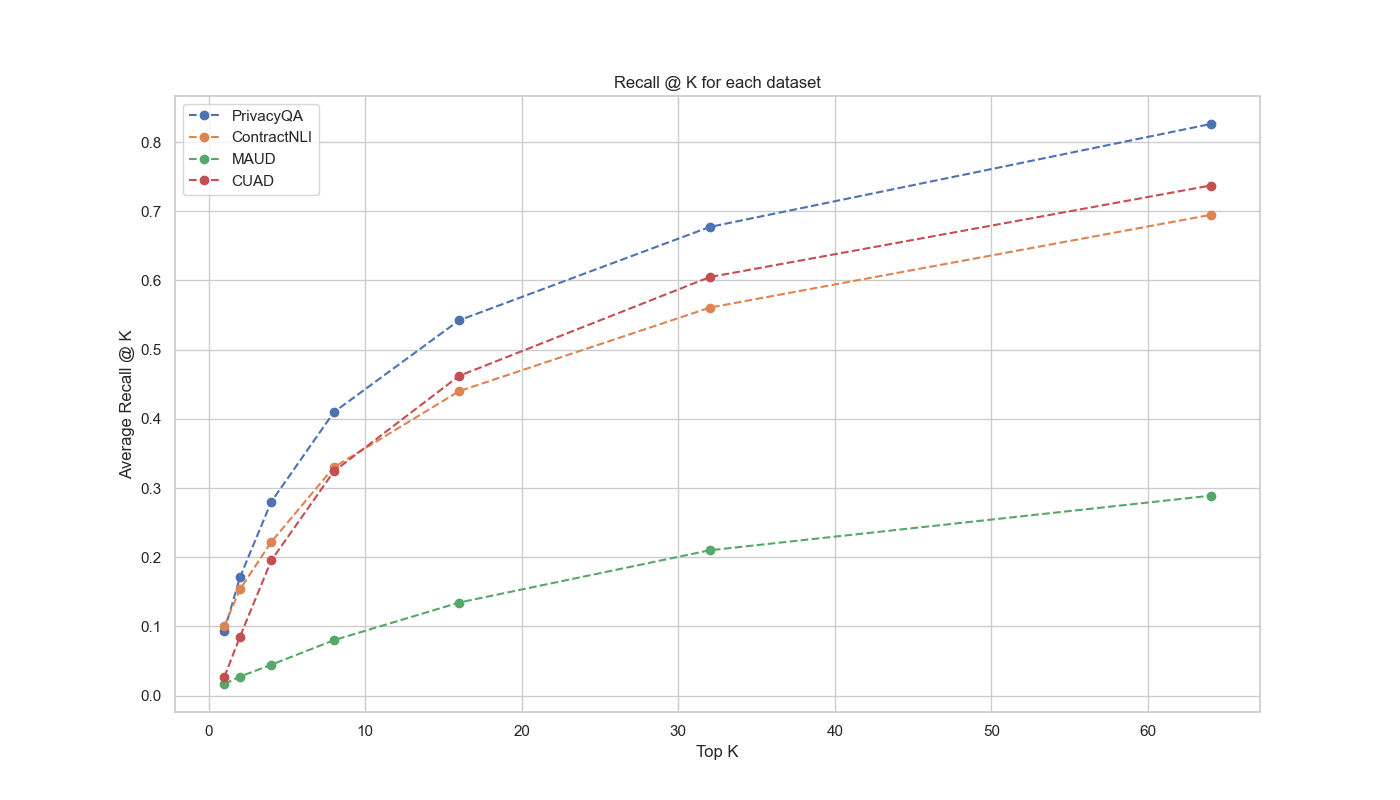}}
\caption{Recall @ $k$ across all datasets}
\label{fig:recall_per_ds}
\end{minipage}\hfill
\begin{minipage}[b]{0.45\textwidth}
        \centering
\centerline{\includegraphics[width=\columnwidth]{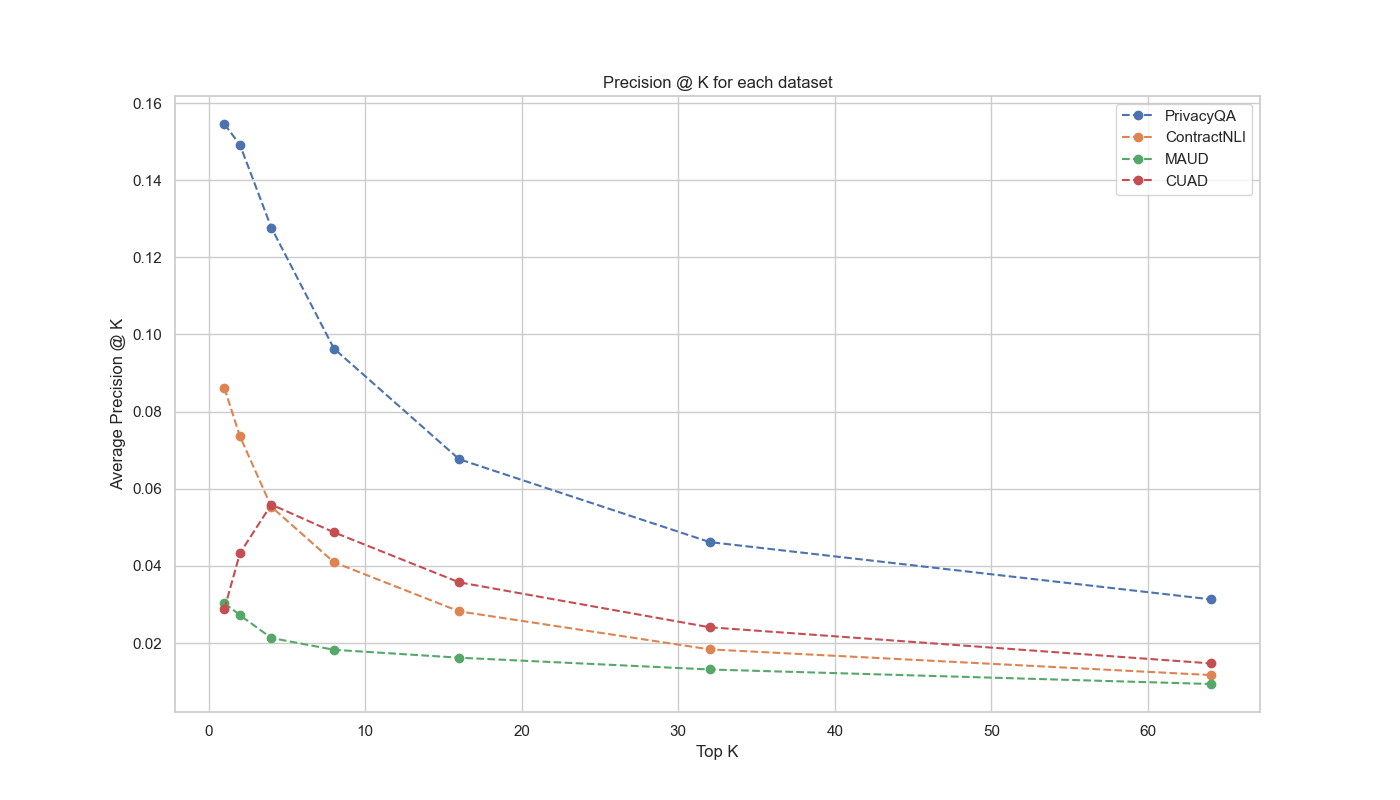}}
\caption{Precision @ $k$ across all datasets}
\label{fig:prec_per_ds}
    \end{minipage}
\vskip -0.2in
\end{figure}

Additionally to the standard evaluation conducted, we also aggregated Recall@$k$ and Precision@$k$ across all four of the different methods experimented with for each of the four datasets. The goal was to assess the relative difficulty of achieving high scores on each dataset, as shown in Figures \ref{fig:recall_per_ds} and \ref{fig:prec_per_ds}.

We analyzed the performance of all methods across the four datasets and found that, as expected, the PrivacyQA dataset is the easiest benchmark. It consistently yielded the highest scores across all methods, indicating that this dataset was easier for the models to interpret. Specifically, on the RCTS with No reranker evaluation, the highest Precision@1 was achieved with 14.38\%, and Recall@64 reached 84.19\%. This high performance can be attributed to the nature of the PrivacyQA dataset, which features questions about private company policies from non-lawyers, making the text more straightforward and less complex.

In contrast, the MAUD dataset proved to be the most challenging. Precision@1 was only 2.65\%, and Recall@64 reached just 28.28\%. The complexity and specialization of the MAUD dataset, which includes highly technical legal jargon, make it particularly difficult for models to retrieve relevant information accurately. The poor performance of the Cohere Reranker on this dataset, and across the board, highlights the limitations of using general-purpose models on specialized legal text.

The ContractNLI and CUAD datasets also presented unique challenges, though they were not as difficult as MAUD. These results suggest that while these datasets are challenging, they are not as inaccessible to retrieval models as MAUD, likely due to their somewhat less specialized language.

\subsection{Future Work}

These results highlight the need for more specialized and challenging legal benchmarks to effectively assess the quality of retrieval systems.
Moreover, the consistently poor performance of the Cohere Reranker, particularly on the more challenging datasets like MAUD, indicates that there is significant room for improvement in reranking models. Future research could explore fine-tuning or developing new rerankers specifically designed for legal text, potentially incorporating more legal-specific features or training on larger, more diverse legal corpora.

% Note use of \abovespace and \belowspace to get reasonable spacing
% above and below tabular lines.

\section{Conclusion}

In conclusion, this paper introduces LegalBench-RAG, the first benchmark specifically designed to evaluate the retrieval component of Retrieval-Augmented Generation (RAG) systems in the legal domain. By leveraging existing expert-annotated legal datasets and meticulously mapping question-answer pairs back to their original contexts, LegalBench-RAG provides a robust framework for assessing retrieval precision and recall. The creation of both LegalBench-RAG and its lightweight counterpart, LegalBench-RAG-mini, addresses a critical gap in the evaluation of RAG systems, particularly in contexts where legal accuracy is paramount.

Through a series of experiments, we demonstrated the effectiveness of different chunking strategies and post-processing methods, revealing that advanced chunking techniques like the Recursive Text Character Splitter (RTCS) significantly enhance retrieval performance. Interestingly, the use of general-purpose rerankers, such as Cohere's model, showed limitations, highlighting the need for domain-specific tools in legal AI applications.

The results underscore the importance of specialized benchmarks that can accurately capture the nuances of legal text retrieval. LegalBench-RAG not only facilitates a more granular evaluation of retrieval mechanisms but also provides a foundation for further advancements in the development of RAG systems tailored to the legal field. As legal AI continues to evolve, the availability of such targeted benchmarks will be crucial for driving innovation and ensuring the reliability of AI-driven legal tools.

% In the unusual situation where you want a paper to appear in the
% references without citing it in the main text, use \nocite

\bibliography{./legalbenchrag}
\bibliographystyle{plainnat}

%%%%%%%%%%%%%%%%%%%%%%%%%%%%%%%%%%%%%%%%%%%%%%%%%%%%%%%%%%%%%%%%%%%%%%%%%%%%%%%
%%%%%%%%%%%%%%%%%%%%%%%%%%%%%%%%%%%%%%%%%%%%%%%%%%%%%%%%%%%%%%%%%%%%%%%%%%%%%%%
% APPENDIX
%%%%%%%%%%%%%%%%%%%%%%%%%%%%%%%%%%%%%%%%%%%%%%%%%%%%%%%%%%%%%%%%%%%%%%%%%%%%%%%
%%%%%%%%%%%%%%%%%%%%%%%%%%%%%%%%%%%%%%%%%%%%%%%%%%%%%%%%%%%%%%%%%%%%%%%%%%%%%%%

\onecolumn
\appendix
\newpage
\section{Appendix A}
\label{sec:appendixA}
\lstdefinelanguage{json}{
  basicstyle=\ttfamily\small,
  numbers=left,
  numberstyle=\tiny,
  stepnumber=1,
  numbersep=8pt,
  showstringspaces=false,
  breaklines=true,
  frame=lines,
  backgroundcolor=\color{gray!10},
  literate=
   *{0}{{{\color{blue}0}}}{1}
    {1}{{{\color{blue}1}}}{1}
    {2}{{{\color{blue}2}}}{1}
    {3}{{{\color{blue}3}}}{1}
    {4}{{{\color{blue}4}}}{1}
    {5}{{{\color{blue}5}}}{1}
    {6}{{{\color{blue}6}}}{1}
    {7}{{{\color{blue}7}}}{1}
    {8}{{{\color{blue}8}}}{1}
    {9}{{{\color{blue}9}}}{1}
    {:}{{{\color{red}:}}}{1}
    {,}{{{\color{red},}}}{1}
    {\{}{{{\color{orange}\{}}}{1}
    {\}}{{{\color{orange}\}}}}{1}
    {[}{{{\color{orange}[}}}{1}
    {]}{{{\color{orange}]}}}{1},
}
\textit{Note: Ellipses (...) were added to replace large regions of text, for readability purposes.} 
\begin{lstlisting}[language=json,firstnumber=1]
{
    "tests": [
        {
            "query": "Consider the Non-Disclosure Agreement (...) to the Confidential Information?",
            "snippets": [
                {
                    "file_path": "contractnli/CopAcc_NDA-and-ToP-Mentors_2.0_2017.txt",
                    "span": [
                        11461,
                        11963
                    ],
                    "answer": "Any and all proprietary rights, including but not limited to rights to and in inventions(...) with the Copernicus Accelerator 2017."
                }
            ]
        },
        ...
        {
            "query": "Consider the Non-Disclosure Agreement (...) include technical information?",
            "snippets": [
                {
                    "file_path": "contractnli/CopAcc_NDA-and-ToP-Mentors_2.0_2017.txt",
                    "span": [
                        7752,
                        8016
                    ],
                    "answer": "Confidential Information means any Idea disclosed to Mentor, (...) and/or its description and any conclusions. "
                }
            ]
}
\end{lstlisting}
\label{json:data}
\centering
\textit{Appendix A:} Extract of the QA pairs in LegalBench-RAG

%%%%%%%%%%%%%%%%%%%%%%%%%%%%%%%%%%%%%%%%%%%%%%%%%%%%%%%%%%%%%%%%%%%%%%%%%%%%%%%
%%%%%%%%%%%%%%%%%%%%%%%%%%%%%%%%%%%%%%%%%%%%%%%%%%%%%%%%%%%%%%%%%%%%%%%%%%%%%%%

\end{document}